%% file: main.tex
\documentclass{IEEEtran}
\input{misc/customize.tex}

\def\BibTeX{{\rm B\kern-.05em{\sc i\kern-.025em b}\kern-.08em
    T\kern-.1667em\lower.7ex\hbox{E}\kern-.125emX}}
\begin{document}

\title{Extractive summarization\\on a CMOS Ising machine}
\author{Ziqing Zeng, Abhimanyu Kumar, Ahmet Efe, Ruihong Yin,\\Chris H. Kim, Ulya R. Karpuzcu, and Sachin S. Sapatnekar\\
University of Minnesota, Minneapolis, MN 55455, USA.}

\maketitle

\begin{abstract}
Extractive summarization~(ES) aims to generate a concise summary by selecting a subset of sentences from a document while maximizing relevance and minimizing redundancy. Although modern ES systems achieve high accuracy using powerful neural models, their deployment typically relies on CPU or GPU infrastructures that are energy-intensive and poorly suited for real-time inference in resource-constrained environments. In this work, we explore the feasibility of implementing McDonald‐style extractive summarization on a low-power CMOS coupled oscillator-based Ising machine (COBI) that supports integer-valued, all-to-all spin couplings. We first propose a hardware-aware Ising formulation that reduces the scale imbalance between local fields and coupling terms, thereby improving robustness to coefficient quantization: this method can be applied to any problem formulation that requires $k$ of $n$ variables to be chosen. We then develop a complete ES pipeline including (i) stochastic rounding and iterative refinement to compensate for precision loss, and (ii) a decomposition strategy that partitions a large ES problem into smaller Ising subproblems that can be efficiently solved on COBI and later combined. Experimental results on the CNN/DailyMail dataset show that our pipeline can produce high-quality summaries using only integer-coupled Ising hardware with limited precision. COBI achieves 3.1--4.3$\times$ speedups in runtime compared to a brute-force baseline, with performance comparable to software-based Tabu search. In terms of energy, COBI delivers three orders of magnitude reduction compared to the brute-force approach and 2.5 orders of magnitude reduction compared to Tabu search, maintaining competitive summary quality. These results highlight the potential of deploying CMOS Ising solvers for real-time, low-energy text summarization on edge devices.
\end{abstract}

\input{sec/1_Intro}

\input{sec/2_Background}
\input{sec/3_Formulation}
\input{sec/4_Workflow}
\input{sec/5_Experiment}
\input{sec/6_Conclusion}
\bibliographystyle{misc/IEEEtran}
\bibliography{misc/IEEEabrv, bib/main}
\end{document}

%% file: misc/customize.tex
\usepackage{cite}
\usepackage{amsmath,amssymb,amsfonts}
\usepackage{graphicx}
\usepackage{textcomp}

\usepackage{multirow}
\usepackage{multicol}
\usepackage{color}
\usepackage{url}
\usepackage{array}
\usepackage{algpseudocode}
\usepackage{float}
\usepackage{enumitem}
\usepackage{makecell}

\usepackage{mathtools}

\usepackage{cases}
\usepackage{subcaption}

\newcommand{\ignore}[1]{}

\usepackage[normalem]{ulem}

%% file: sec/1_Intro.tex
\section{Introduction}
\label{sec:ES_Intro}
Extractive summarization~(ES) is the task of selecting a subset of sentences from a document or a collection of documents to generate a concise and informative summary. ES plays a critical role in many real-world applications, such as generating news digests, legal or medical briefings and meeting notes~\cite{filippova2009company,bhattacharya2021incorporating}. As the volume of digital content continues to grow, there is an increasing demand for automated summarization tools that can operate efficiently and effectively across diverse domains. Despite its utility, ES is a computationally challenging task. 
There are multiple approaches to solving ES~\cite{mcdonald2007study, xu2020discourse, zhong2020extractive}. One typical formulation~\cite{mcdonald2007study} casts it as a global optimization problem that selects a subset of textual units (e.g., sentences) that maximizes overall informativeness (centrality relevance) while minimizing overlap in content (redundancy), subject to a total length budget in a number of sentences. Even under simplified assumptions such as pairwise redundancy, this formulation is NP-hard~\cite{mcdonald2007study}. As a result, various approximate inference techniques, such as greedy search and dynamic programming heuristics, have been explored.

Under the simplified pairwise redundancy assumption, McDonald-style ES~\cite{mcdonald2007study} can be formulated as an Integer Linear Programming~(ILP) problem, enabling formal optimization under hard constraints. The ILP model can be mapped into a Quadratic Unconstrained Binary Optimization (QUBO) problem, or equivalently into an Ising model, by introducing penalty coefficients to enforce the constraints~\cite{niroula2022constrained}. This representation allows the use of specialized Ising solvers. However, solving the ILP formulation with high precision is computationally intensive. Software-based solvers such as Tabu search and the Quantum Approximate Optimization Algorithm~(QAOA) often require substantial runtime, especially as the size, resolution, or density of the problem increases~\cite{misevicius2005tabu,farhi2014quantum,akshay2020reachability}.  
Quantum-based methods such as the D-Wave quantum annealer can leverage quantum tunneling to accelerate combinatorial optimization, but they require cryogenically cooled superconducting hardware and offer limited qubit connectivity~\cite{hamerly2019experimental,tasseff2024emerging}. These limitations result in significant power consumption and degraded performance when applied to dense problems such as extractive summarization.

In this work, we propose leveraging hardware-aware approximations to enable the simplified ES formulation~\cite{mcdonald2007study} on the Coupled Oscillator-based Ising~(COBI) solver~\cite{Lo2023}. The recently demonstrated 48-node all-to-all CMOS-based COBI solver~\cite{cilasun2025coupled} offers ultra-fast convergence, low power consumption~(approximately 24 mW), and support for 5-bit integer coupling weights. As a physics-based analog system, COBI solves Ising problems by allowing a network of coupled ring oscillators to naturally evolve into low-energy states, thereby eliminating the need for conventional iterative search algorithms and digital control overhead. However, its limited precision and number of spins pose significant challenges when attempting to directly implement high-resolution formulations derived from Bidirectional Encoder Representations from Transformers~(BERT) embeddings. 

To address these hardware constraints, we develop a complete workflow that begins with the ILP-based ES formulation and applies a series of hardware-aware transformations. Specifically, our approach includes: 
\begin{enumerate}
\item[(1)] designing an improved Ising formulation based on the ILP model~\cite{niroula2022constrained}, which balances the constraints and objective; 
\item[(2)] proposing an approach for shifting the linear coefficients of the Ising formulation to reduce imbalances between linear and quadratic coefficients: this approach is general enough for application to any QUBO/Ising formulation that requires $k$ of $n$ variables to be selected, such as~\cite{feld2019hybrid,zeng2025im} and the traveling saleman problem in~\cite{lucas2014ising};
\item[(3)] proposing stochastic rounding~\cite{croci2022stochastic} to iteratively quantize the floating-point sentence relevance scores ($\mu_i$) and pairwise redundancy terms ($\beta_{ij}$) into the integer range supported by COBI, leveraging its high-speed execution to compensate for reduced precision; and 
\item[(4)] introducing a decomposition strategy that partitions the summarization problem into smaller subproblems, solves them separately on hardware, and integrates their outputs into a coherent, high-quality summary.
\end{enumerate}

Our results demonstrate that extractive summaries can be produced under tight precision and size constraints by integrating algorithmic approximations with hardware-specific design principles. In particular, the COBI-based ES solver delivers consistent reductions in time-to-solution~(TTS) and achieves orders-of-magnitude improvements in energy-to-solution~(ETS) compared to software-based Tabu Search and brute-force baselines, demonstrating the practical value of analog CMOS Ising hardware for fast and energy-efficient ES. These findings open promising opportunities for deploying real-time, low-power summarization engines in edge devices, embedded systems, and content-aware communication platforms.

%% file: sec/2_Background.tex
\section{Background}
\label{sec:ES_Background}

\subsection{Extractive summarization}
\label{ssec:ES_Background_es}

ES seeks to generate a concise summary by selecting a subset of sentences from a source document, preserving key information content while avoiding redundancy. Unlike abstractive summarization, which generates novel sentences, ES draws directly from the input text, making it more interpretable and better suited to domains requiring high factual accuracy, such as legal and medical summarization~\cite{filippova2009company,bhattacharya2021incorporating}.

Formally, let $\mathcal{S} = \{s_1, s_2, \dots, s_n\}$ denote the $N$ candidate sentences in the input document.  For this set, we associate scores $\mu_i$ and $\beta_{ij}$, where $\mu_i$ is the relevance score of sentence $s_i$, capturing its informativeness, and $\beta_{ij}$ is the redundancy penalty between sentences $s_i$ and $s_j$.  In practice, semantic features extracted from transformer-based language models such as Sentence-BERT~\cite{reimers2019sentence} can be used to determine these scores.  Let $\mathbf{e}_i$ be the embedding of sentence $s_i$, and let $\bar{\mathbf{e}}_{\text{doc}}$ denote the mean embedding of the full document, computed from the BERT layer~\cite{devlin2019bert} and pooling~\cite{reimers2019sentence}. Accordingly, $\mu_i$ and $\beta_{ij}$ can be defined as
\begin{align}
    \mu_i &= \cos(\mathbf{e}_i, \bar{\mathbf{e}}_{\text{doc}})\\
    \beta_{ij} &= \cos(\mathbf{e}_i, \mathbf{e}_j),
\end{align}
where $\cos(\cdot, \cdot)$ denotes the cosine similarity function.  

A well-established formulation by McDonald~\cite{mcdonald2007study} models ES as a constrained optimization problem. Sentences are scored based on their relevance to the document and pairwise redundancy with other selected sentences. The task is to select a subset of sentences that maximizes a linear objective function combining relevance and redundancy, subject to a length constraint (e.g., number of sentences or total tokens). This formulation is NP-hard even under simplified assumptions, motivating approximate solutions. Defining a binary variable $x_i \in \{0, 1\}$ to indicate whether sentence $s_i$ is selected into the summary~(1 for selected, 0 for not selected), the extractive summarization task can be framed as the following optimization problem:
\begin{equation}
\begin{aligned}
\max_{x \in \{0,1\}^n} \quad & \sum_{i=0}^{N-1} \mu_i x_i - \lambda\sum_{i\neq j} \beta_{ij} x_i x_j \\
&\text{subject to} \quad  \sum_{i=0}^{N-1} x_i = M,
\end{aligned}
\label{eq:es_opt_mu_beta}
\end{equation}
where $M$ is the summary length budget, typically expressed as the number of sentences in the summary. In the objective function, the relevance score $\mu_i$ encourages the selection of sentences that are semantically close to the overall document, while the redundancy score $\beta_{ij}$ penalizes the joint inclusion of semantically similar sentences. Therefore, the optimization seeks to select sentences that maximize content coverage (through high $\mu_i$ values) while minimizing overlap (penalizing high $\beta_{ij}$ values between selected pairs). The constraint enforces a limit on the total summary length. In practice, since every sentence in a paragraph tends to have some correlation with every other sentence, $\beta_{ij} \neq 0 \; \forall \; i,j$.

An illustrative example applying the McDonald model to generate a six-sentence summary is provided in the Supplementary Materials. The example demonstrates that the model effectively identifies sentences containing the most salient information. However, because the formulation optimizes solely for centrality and redundancy, linguistic coherence and fluency are not explicitly considered.

Overall, the formulation in~\eqref{eq:es_opt_mu_beta} provides a principled and flexible framework for extractive summarization. Nevertheless, its exact solution is NP-hard~\cite{mcdonald2007study}, motivating the use of heuristic methods or alternative optimization backends, such as QUBO or Ising solvers, to approximate high-quality solutions efficiently.

\subsection{QUBO and Ising solvers}
\label{ssec:ES_Background_ising}

The Ising model, originally developed in statistical physics to describe spin interactions in magnetic materials, has become a powerful framework for solving NP-hard combinatorial optimization problems. In computational settings, an optimization problem is encoded as an Ising Hamiltonian of the form:
\begin{equation}
\label{eq:ising_std}
\min_{s \in \{-1,+1\}^n} \mathcal{H}(s) = \sum_i h_i s_i + \sum_{i \ne j} J_{ij} s_i s_j,
\end{equation}
where $s_i \in {-1, +1}$ are spin variables, $h_i$ represents the local field on spin $i$, and $J_{ij}$ denotes the interaction strength between spins $i$ and $j$. This formulation captures both unary and pairwise contributions to the energy of a spin configuration, and the goal is to find a configuration $s$ that minimizes $\mathcal{H}(s)$.

The Ising model is mathematically equivalent to the QUBO formulation, which operates on binary variables $x_i \in {0, 1}$. A general QUBO problem is expressed as:
\begin{equation}
\label{eq:qubo_general}
\min_{x \in \{0,1\}^n}\mathcal{H}(x) = \sum_i Q_{ii} x_i + \sum_{i \neq j} Q_{ij} x_i x_j,
\end{equation}
where $Q_{ii}$ and $Q_{ij}$ are real-valued linear and quadratic coefficients. The equivalence between QUBO and Ising is established by the change of variables $x_i = \frac{1 + s_i}{2}$, mapping binary variables to spin variables. Substituting this relation into Eq.~\eqref{eq:qubo_general} yields the Ising Hamiltonian with parameters:
\begin{equation}
\label{eq:qubo_to_ising}
h_i = \frac{1}{2} Q_{ii} + \frac{1}{4} \sum_{i \ne j} Q_{ij},\quad J_{ij} = \frac{1}{4} Q_{ij}
\end{equation}
This equivalence enables the transformation of many classical optimization problems, such as Max-cut, Max-SAT, and certain integer programming formulations, into Ising form~\cite{lucas2014ising}, making them amenable to solution via analog Ising machines or quantum annealing hardware.

The connectivity topology of an Ising machine is a key factor in its ability to embed and solve different problem classes. For instance, the Chimera topology used in D-Wave 2000Q consists of a 2D grid of sparse $K_{4,4}$ bipartite cells, and has a limited connectivity to 6 neighbors~\cite{willsch2022benchmarking}. The Pegasus topology, introduced in D-Wave Advantage, increases each connectivity of each qubit to its 15 neighbors, enabling more compact embeddings of denser graphs. The Zephyr topology, adopted by D-Wave Advantage2, further improves local connectivity to 20 and embedding efficiency~\cite{pelofske2023comparing}. In contrast to these sparsely structured topologies, all-to-all connectivity—where every spin can couple to every other—represents an ideal setting for fully connected optimization problems but is challenging to achieve in practice. This all-to-all coupling capability is particularly important for the ES problem, where, as we will show, $J_{ij}$ tracks $\beta_{ij}$; therefore, $J_{ij} \neq 0 \; \forall \; i,j$, resulting in all-to-all coupling.

CMOS-based analog Ising machines such as the COBI solver chip~\cite{Lo2023,cilasun2025coupled} overcome the connectivity limitation of prior hardware approaches by directly implementing an all-to-all connected architecture on a CMOS ring oscillator array. The 59-spin COBI system allows every spin to interact with all others via programmable integer-valued coupling weights $h_i, J_{ij} \in [-14, +14]$, enabling dense QUBO/Ising problems to be solved without the need for topological embedding. The system leverages the phase dynamics of coupled ring oscillators to solve the Ising problem with ultra-fast convergence and extremely low power consumption~(i.e., 25 mW), representing orders of magnitude lower energy usage than typical quantum annealers. Furthermore, COBI operates at room temperature, in stark contrast to quantum annealing systems like D-Wave, which require cryogenic cooling to milliKelvin temperatures. COBI removes the need for complex and costly cooling infrastructure and significantly enhances the practicality and portability of the system.

Despite these advantages, the COBI hardware introduces several practical constraints: it supports only a limited number of spins, requires integer-valued coefficients, and lacks floating-point precision. To address these constraints, various preprocessing techniques, such as weight scaling, spin merging, and spin pruning, are applied to transform high-precision Ising problems into a hardware-compatible format without sacrificing essential problem structure~\cite{cilasun20243sat}.

%% file: sec/3_Formulation.tex
\section{Ising formulation for extractive summarization}
\label{sec:ES_formulation}
In Section~\ref{sec:ES_Background}, an optimization formulation of ES is summarized in~\eqref{eq:es_opt_mu_beta} as an optimization problem with binary variables. The optimization objective has linear and quadratic terms based on the binary variables, and there is a constraint on the variable. To solve this problem with an Ising solver. It should be transferred into a Quadratic Unconstrained Binary Optimization~(QUBO) formulation by introducing a penalty term to enforce the constraint~\cite {niroula2022constrained}. The penalty term is minimized when the constraint is satisfied that exactly $M$ sentences are selected in the summary. After adding a penalty coefficient $\Gamma$ as the weight of the penalty term, the formulation in~\eqref{eq:es_opt_mu_beta} is transferred into 
\begin{equation}
\label{eq:es_opt}
\max_{x \in \{0,1\}^N}\mathcal{H}(x)= \sum_{i=0}^{N-1} \mu_i x_i - \lambda\sum_{i\neq j} \beta_{ij} x_i x_j -\Gamma(\sum_{i=0}^{N-1}x_{i}-M)^2
\end{equation}
The above expression can be further transferred into a standard QUBO formulation by reversing the objective into a minimization and ignoring the constant terms.
\begin{equation}
\label{eq:es_qubo}
\min_{x \in \{0,1\}^N}\mathcal{H}(x)= \sum_{i=0}^{N-1} (-\mu_i-2\Gamma M+\Gamma)x_i + \sum_{i\neq j} (\lambda\beta_{ij}+\Gamma) x_i x_j
\end{equation}

To deploy the ES objective on Ising hardware, we must convert the QUBO formulation in Eq.~\eqref{eq:es_qubo} into an equivalent Ising form. 
Applying QUBO-to-Ising transformation, as introduced in Section~\ref{ssec:ES_Background_ising}, on the ES-specific QUBO in Eq.~\eqref{eq:es_qubo}, the corresponding Ising formulation is:
\begin{equation}
\label{eq:es_ising}
\begin{aligned}
\min_{s \in \{-1,+1\}^n} \mathcal{H}(s) = \sum_i h_i s_i + \sum_{i \ne j} J_{ij} s_i s_j,\\
\text{where} \quad
\begin{aligned}
h_i =& \frac{1}{2} \left(-\mu_i - 2\Gamma M + \Gamma\right) \\
&+ \frac{1}{4} \sum_{j \ne i} (\lambda \beta_{ij} + \Gamma), \\
J_{ij} =& \frac{1}{4} (\lambda \beta_{ij} + \Gamma)
\end{aligned}
\end{aligned}
\end{equation}

\subsection{Challenges in solving the Ising formulation}
\label{ssec:ES_formulation_challenge}

\noindent
Based on the Ising formulation in~\eqref{eq:es_ising}, we use Sentence-BERT~\cite{reimers2019sentence} to compute the relevance scores $\mu_i$ and redundancy penalties $\beta_{ij}$ for each sentence pair and generate the corresponding Ising formulation for ES. However, two key properties of this formulation introduce challenges for hardware implementation on COBI, which can accommodate integer coupling coefficients with up to 5-bit precision: (1)~The dense Ising formulation results in subproblems with a very wide range of Ising coupling coefficients (2)~The resulting coefficients are expressed in floating-point precision.

Since all centrality scores $\mu_i$ and redundancy scores $\beta_{ij}$ are nonzero by construction, the resulting Ising coefficients $h_i$ and $J_{ij}$ are also nonzero. Consequently, every benchmark yields a fully dense Ising model in which \emph{all} spins and couplers are active. Deploying such dense formulations on sparsely connected hardware architectures (e.g., Chimera~\cite{katzgraber2014glassy} or Pegasus~\cite{dattani2019pegasus}) requires complex minor embeddings, which incur both mapping overhead and additional physical qubits~\cite{pelofske2023comparing}. In contrast, COBI~\cite{Lo2023}, an all-to-all CMOS-based Ising solver, natively supports full spin connectivity and is therefore well suited for dense problems such as ES. However, COBI also imposes important constraints: it supports only integer-valued coupling weights within a restricted range and provides a limited number of physical spins. As a result, the precision of the coefficients and the size of the Ising formulation must both be reduced in order to map onto COBI.

A second major challenge arises from the discrepancy between the value ranges of $h_i$ and $J_{ij}$. Over the 20-sentence benchmarks from the CNN/DailyMail dataset~(see Supplementary Materials), the $h_i$ values are centered around $\sim3.85$, while the $J_{ij}$ values are centered around $\sim0.52$, i.e., almost an order of magnitude smaller. Nevertheless, the cumulative term $\sum_j J_{ij}$ is comparable in magnitude to $h_i$, indicating that both sets of coefficients make significant contributions to the Hamiltonian. Thus, simply truncating or ignoring the smaller coefficients would severely distort the energy landscape.

Following the preprocessing techniques described in prior work~\cite{cilasun20243sat}, one might consider scaling or truncating the coefficients to fit the integer range of the hardware. However, neither approach is effective in this setting. Scaling all coefficients so that $J_{ij}$ fits within the $[-14,14]$ integer range of COBI would result in $|h_i|$ values exceeding $100$, requiring severe truncation that would distort the formulation. Conversely, scaling $h_i$ into $[-14,14]$ would place all $J_{ij}$ values in a narrow range (e.g., $[1.8,1.98]$), which would then be rounded to the same integer value and eliminate variability across interactions. In both cases, the resulting Hamiltonian deviates substantially from the original formulation. Consequently, conventional preprocessing techniques such as uniform scaling or truncation are inadequate for preserving formulation fidelity while satisfying the hardware constraints of COBI.

\subsection{Improved formulation under limited bit precision for the coupling coefficient}
\label{ssec:ES_Formulation_improved}

To address the challenges discussed above, we propose a modified Ising formulation designed to reduce the discrepancy between the distributions of the local field terms $h_i$ and the pairwise coupling terms $J_{ij}$. This rebalancing improves compatibility with COBI hardware by aligning coefficient scales more effectively with its limited integer coupling range, thereby avoiding excessive quantization error.

Recall from the original constrained optimization problem in Eq.~\eqref{eq:es_opt_mu_beta} that any feasible solution must satisfy the cardinality constraint $\sum_i x_i = M$, where $M$ is a fixed constant representing the number of sentences selected. This constraint is preserved in the QUBO formulation, Eq.~\eqref{eq:es_qubo_opt}, through a penalty term weighted by the coefficient $\Gamma$. Furthermore, since adding a constant to the objective function does not alter the optimal solution, we can safely augment the objective by a linear bias term $\mu_b \sum_i x_i$, where $\mu_b$ is a tunable scalar. This enables us to shift the linear coefficients without changing the solution set.

Incorporating this bias term, the adjusted QUBO formulation becomes:
\begin{equation}
\label{eq:es_qubo_opt}
\begin{aligned}
\max_{x \in {0,1}^N} \mathcal{H}(x) = & \sum_{i=0}^{N-1} (\mu_i + \mu_b) x_i - \lambda \sum_{i \ne j} \beta_{ij} x_i x_j \\
& -\Gamma \left( \sum_{i=0}^{N-1} x_i - M \right)^2.
\end{aligned}
\end{equation}
This adjustment increases the relative weight of the linear terms and narrows the scale gap between the linear and quadratic components, enabling more hardware-friendly mapping to integer-valued Ising parameters within the $[-14, +14]$ range of COBI.

We then convert this improved QUBO into an Ising formulation following the standard QUBO-to-Ising transformation described in~\eqref{eq:qubo_to_ising}. The resulting Ising Hamiltonian is:
\begin{equation}
\label{eq:es_ising_final}
\begin{aligned}
\min_{s \in \{-1,+1\}^n} \mathcal{H}(s) &= \sum_i h'_i s_i + \sum_{i \ne j} J'_{ij} s_i s_j, \\
\text{where} \quad
h'_i &= \frac{1}{2} \left(-\mu_i - \mu_b - 2\Gamma M + \Gamma\right)\\
&+ \frac{1}{4} \sum_{j \ne i} (\lambda \beta_{ij} + \Gamma), \\
J'_{ij} &= \frac{1}{4} (\lambda \beta_{ij} + \Gamma)
\end{aligned}
\end{equation}

We define the bias term $\mu_b$ based on the medians of the original coefficient distributions:
\begin{equation}
u_b = 2(\text{median}(h_i) - \text{median}(J_{ij}))
\end{equation}
This choice ensures that the median of $h'_i$ aligns with the median of $J'_{ij}$ for each benchmark instance.

To evaluate the effectiveness of the improved formulation, we apply both the original Ising formulation~\eqref{eq:es_ising} and the improved variant~\eqref{eq:es_ising_final} to the 20 benchmark problems shown. Each benchmark is solved using a Tabu search algorithm~\cite{glover1998tabu} under a range of numerical precision settings: floating-point full precision~(FP), fixed-point formats with 6, 5, and 4 bits, and integer-valued coupling weights constrained to the range $[-14, +14]$, which corresponds to the native resolution of the COBI hardware. Lower-precision formats are simulated by quantizing the original floating-point weights to the desired bit-width. We begin the analysis from 6-bit precision since, for 20/50-sentence paragraphs, fixed-point higher precisions exhibit behavior closely aligned with FP performance. Additional results for 7- and 8-bit configurations are provided in the Supplementary Materials, which are closer to FP.

After solving each instance, we normalize the resulting objective values using the following transformation:
\begin{equation}
\label{eq:normalized_obj}
    \text{Normalized Objective} = \frac{\text{obj} - \text{obj}_{\min}}{\text{obj}_{\max} - \text{obj}_{\min}},
\end{equation}
where $\text{obj}$ denotes the objective value computed from the Ising solver solution, evaluated using the objective function~\eqref{eq:es_opt_mu_beta} under FP. The terms $\text{obj}_{\min}$ and $\text{obj}_{\max}$ represent the minimum and maximum objective values obtained via the commercial optimization tool Gurobi~\cite{gurobi}, and serve as reference bounds for normalization. This scaling maps all objective values to the interval $[0, 1]$, enabling fair and consistent comparison across different formulations and precision settings.

\begin{figure}[t]
\centering
\includegraphics[width=0.7\linewidth]{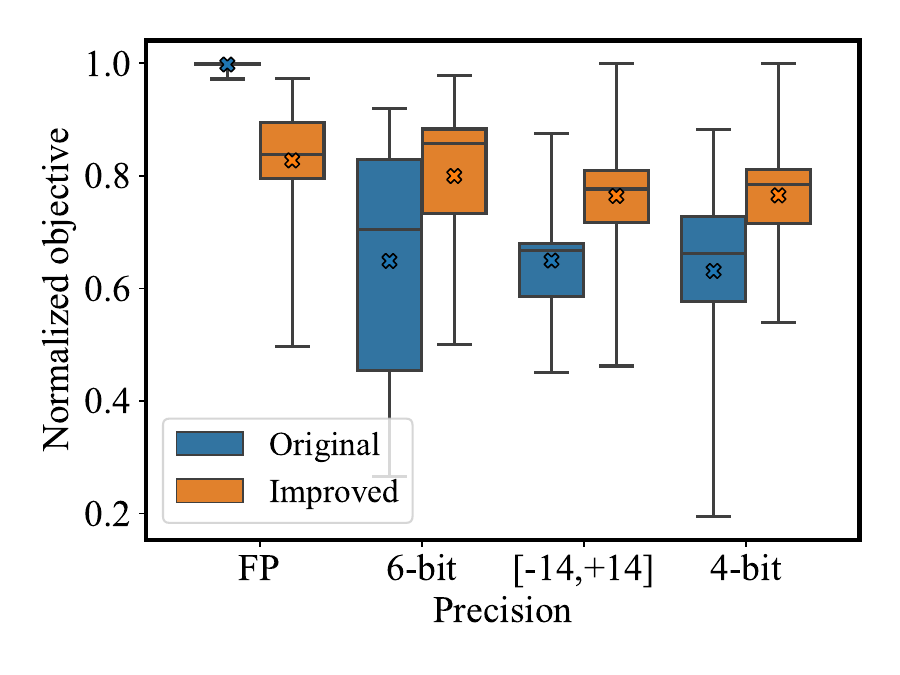}
\vspace{-0.1in}
\caption{Distribution of normalized objective for 20-sentence benchmarks.}
\label{fig:ES_normalized_objs}
\end{figure}

Figure~\ref{fig:ES_normalized_objs} presents the distribution of normalized objectives across all 20 benchmarks for both the original and improved formulations, under original FP and man fixed-bit precisions. In the plot, boxes indicate the 25th, 50th, and 75th percentiles, whiskers represent the minimum and maximum, and the mean is marked with a cross. The original formulation under full-precision arithmetic achieves a normalized objective above 0.99, showing it nearly always recovers the optimal solution. When rounded to 6-bit precision, however, its normalized score drops to 0.66, demonstrating sensitivity to limited numerical precision on an Ising solver. In contrast, the improved formulation attains a slightly lower average of 0.83 under full precision but maintains better performance at 6-bit precision, reaching 0.74. Detailed results for each benchmark are provided in the supplementary materials.

These results illustrate a clear trade-off. Although the improved formulation sacrifices some accuracy at full precision, it significantly enhances robustness under low precision, making it more suitable for deployment on hardware Ising solvers such as COBI. Introducing a bias term into the Ising formulation helps mitigate performance degradation when mapping floating-point weights onto hardware solvers that support only low-precision coupling weights.

%% file: sec/4_Workflow.tex
\section{Workflow exploration of ES based on Simulations}
\label{sec:ES_workflow}
As discussed in Section~\ref{ssec:ES_Formulation_improved}, the improved formulation enhances overall performance.  Figure~\ref{fig:ES_normalized_objs} shows that while this formulation improves the average-case behavior, some benchmarks still have low minimum normalized objectives~(e.g., below 0.5). To address this, we incorporate two strategies into our workflow: iteration with stochastic rounding and decomposition, aiming to further improve the performance.

\subsection{Iterations with stochastic rounding}
\label{ssec:ES_iterations}

The iterative approach specifically targets the degradation in solution quality that occurs when floating-point Ising coefficients are quantized to limited-precision integers. In practice, we observe that the resulting quantized Ising formulations often exhibit multiple optimal solutions, i.e., distinct spin configurations that yield the same minimum energy. Experiments show that a nonnegligible fraction of these quantized formulations admit two or more equivalent optima, as detailed in the Supplementary Materials.

This multiplicity of optima introduces two key challenges. First, while the quantized formulation may admit multiple global optima, many of these lie at a considerable Hamming distance from the true optimal solution of the original floating-point formulation, and only a subset corresponds to high-quality summaries when evaluated under the original objective. Second, even when several configurations yield nearly identical energies in both the quantized and original formulations, the Ising solver may still fail to recover a desirable solution in a single run, leaving the probability of obtaining a high-quality outcome from one hardware execution low.

To mitigate this issue, we adopt an iterative refinement strategy. In each iteration, we apply a rounding scheme to generate a quantized Ising instance, solve it using the COBI solver, and evaluate the resulting solution using the original floating-point objective. After $i$ iterations, the candidate with the highest objective value is selected as the final output. Although the total runtime increases linearly with the number of iterations, each solver invocation is extremely fast on hardware platforms such as COBI. Thus, this approach provides an effective trade-off: a modest increase in runtime yields a substantially higher likelihood of recovering a high-quality solution under limited-precision hardware.

We explore three rounding schemes for generating quantized Ising formulations:
\begin{itemize}
    \item \textbf{Deterministic rounding:} Floating-point coefficients are rounded to their nearest integers. The same quantized formulation is solved multiple times to explore variability in solver outcomes.
    \item \textbf{Stochastic (50/50) rounding:} Each coefficient is independently rounded up or down with \uline{equal probability (50\%)}, creating multiple randomized instances of the problem.
    \item \textbf{Stochastic rounding:} Each coefficient is \uline{probabilistically rounded} based on its fractional part, such that values closer to an integer are more likely to round toward it. This introduces controlled noise while preserving the overall statistical structure of the weights.
\end{itemize}
As a \textbf{baseline}, we bypass the Ising formulation entirely and instead generate summaries by randomly selecting $M$ sentences from the paragraph in each iteration. This serves as a reference point for evaluating the added value of hardware-based optimization.

In our results, \textbf{Number of iterations} is defined as the number of Ising formulations solved by Tabu and COBI. For the random baseline without any Ising solving, \textbf{Number of iterations} is defined as the number of solutions whose objective function value is compared to the optimal solution.

\begin{figure}[t]
\centering
    \begin{subfigure}[b]{0.5\linewidth}
        \centering
        \includegraphics[width=\linewidth]{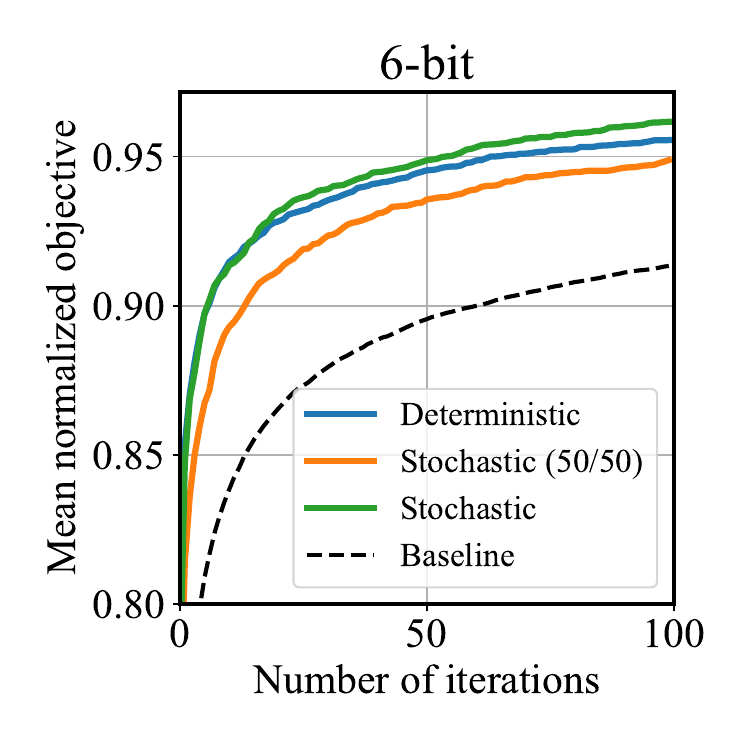}
        \vspace{-0.3in}
        \caption{}
    \end{subfigure}
    \hspace{-0.02\linewidth} 
    \begin{subfigure}[b]{0.5\linewidth}
        \centering
        \includegraphics[width=\linewidth]{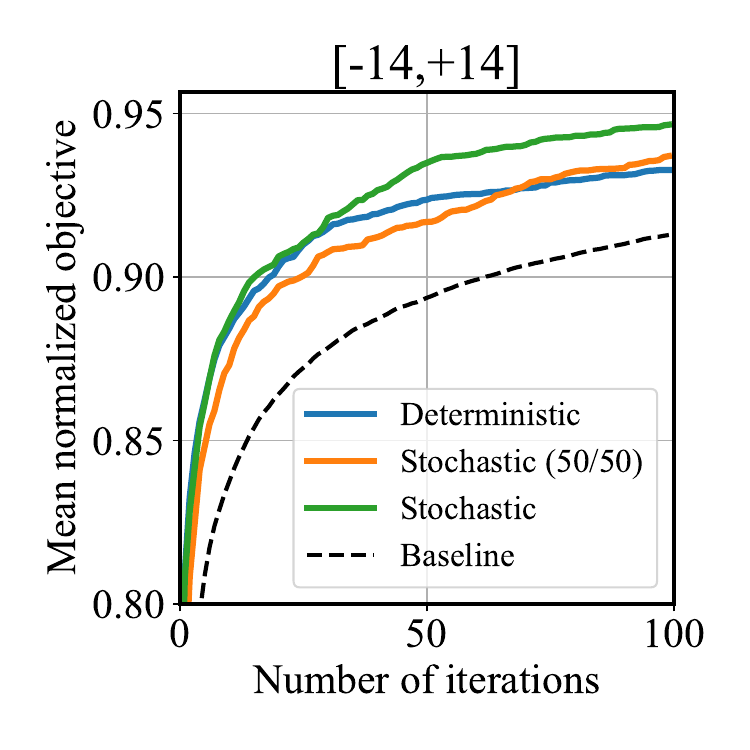}
        \vspace{-0.3in}
        \caption{}
    \end{subfigure}
        \begin{subfigure}[b]{0.5\linewidth}
        \centering
        \includegraphics[width=\linewidth]{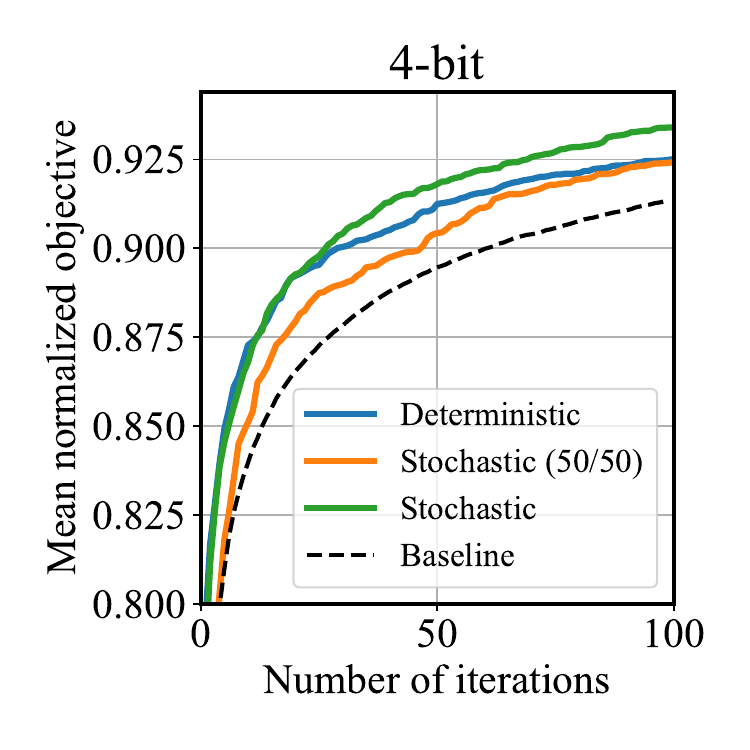} 
        \vspace{-0.3in}
        \caption{}
    \end{subfigure}
\caption{Normalized objective on 20-sentence benchmarks, where (a)--(d) correspond to varying levels of precision.}
\label{fig:ES_iterations_20}
\end{figure}

\begin{figure}[ht]
\centering
    \begin{subfigure}[b]{0.5\linewidth}
        \centering
        \includegraphics[width=\linewidth]{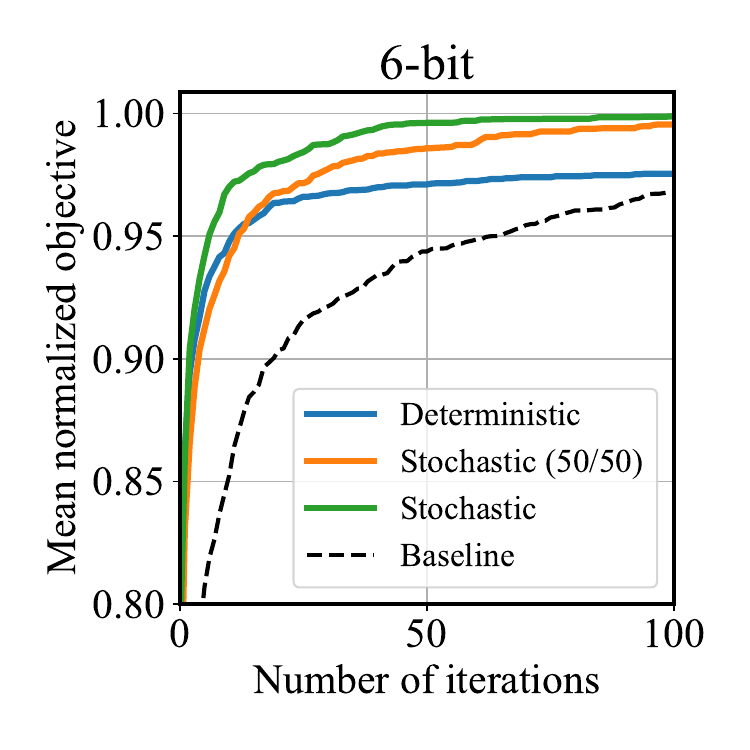}
        \vspace{-0.3in}
        \caption{}
    \end{subfigure}
    \hspace{-0.02\linewidth} 
    \begin{subfigure}[b]{0.5\linewidth}
        \centering
        \includegraphics[width=\linewidth]{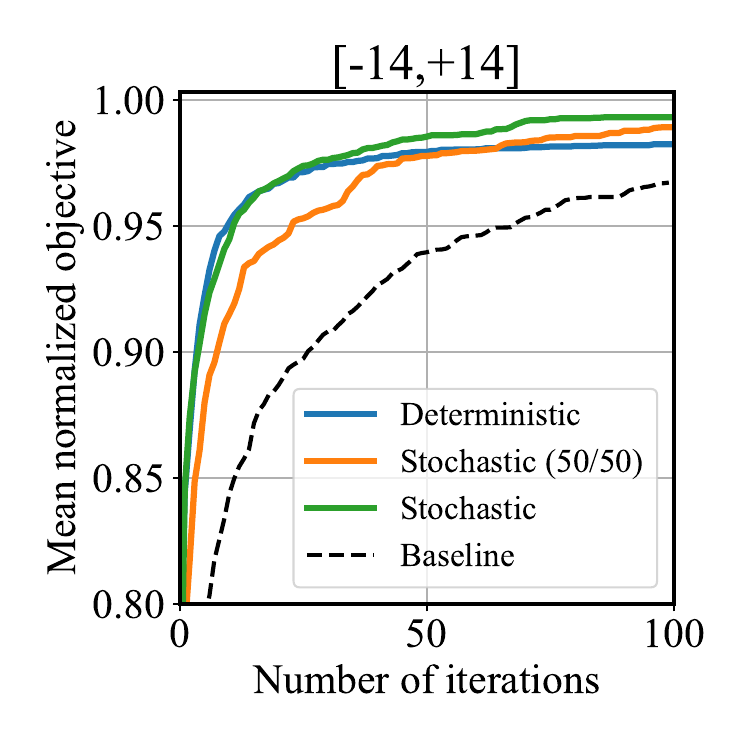}
        \vspace{-0.3in}
        \caption{}
    \end{subfigure}
        \begin{subfigure}[b]{0.5\linewidth}
        \centering
        \includegraphics[width=\linewidth]{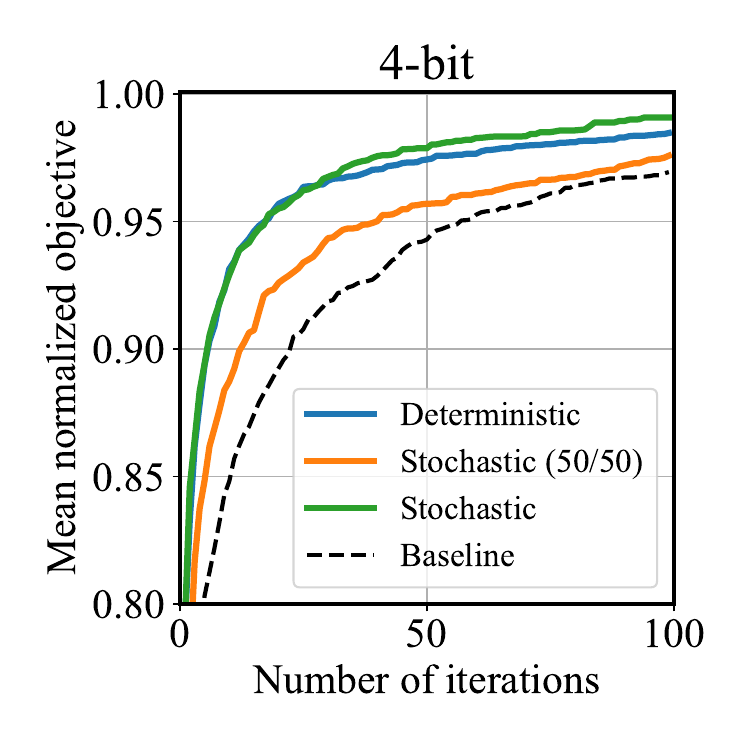} 
        \vspace{-0.3in}
        \caption{}
    \end{subfigure}
\caption{Normalized objective on 10-sentence benchmarks, where (a)--(d) correspond to varying levels of precision.}
\label{fig:ES_iterations_10}
\end{figure}

We evaluate the effectiveness of iterative refinement using three rounding strategies, including deterministic, Stochastic 50-50, and stochastic rounding across 20 benchmarks from the CNN/DailyMail dataset. Each benchmark consists of a paragraph containing at least 20 sentences, from which we construct Ising formulations using the first 20 sentences. For each case, we apply the iterative solver with 1 to 100 iterations, computing the normalized objective value after each iteration according to Eq.~\eqref{eq:normalized_obj}. To account for randomness and ensure statistical robustness, we perform 10 independent runs for each benchmark and report the average normalized objective across all iterations and benchmarks. The results for 20-sentence and 10-sentence benchmarks are presented in Figure~\ref{fig:ES_iterations_20} and  Figure~\ref{fig:ES_iterations_10}.

From Figure~\ref{fig:ES_iterations_20}, we see that increasing the number of iterations consistently boosts the normalized objective across all rounding schemes. Stochastic rounding achieves the highest overall performance for most precision settings. Stochastic 50-50 rounding, however, performs poorly at low precisions (e.g., 4-bit) because its indiscriminate up/down rounding introduces large perturbations to the Ising Hamiltonian, severely degrading solution quality. Deterministic rounding converges quickly and saturates after a few iterations, since it always produces the same quantized Hamiltonian and thus cannot explore alternative solution spaces. At higher precisions (6/7/8-bit), all three rounding schemes converge to similar results, as the effect of rounding becomes negligible. In Figure~\ref{fig:ES_iterations_10}, the performance gains from stochastic rounding vary with precision, but the overall trend remains: Stochastic rounding consistently benefits from iterative refinement and delivers the best results across settings.

Overall, due to the superior and consistent performance of stochastic rounding, we select it as our default rounding strategy for all subsequent experiments. Its ability to balance diversity and structure during quantization makes it especially effective for mitigating the limitations of low-precision hardware solvers.

\subsection{Decomposition}
\label{ssec:ES_decomposition}
While optimizing the formulation, we observe that the distributions of the local fields $h_i$ and pairwise coupling weights $J_{ij}$ in the resulting Ising models are problem-dependent, even when the benchmark length is held constant. Furthermore, as shown in Eq.~\eqref{eq:es_ising_final}, both $h_i$ and $J_{ij}$ increase with the total number of input sentences $N$ and the desired summary length $M$. When the ES task is directly encoded into an Ising formulation, these values are rigidly defined by the structure of the input and the task specification, offering little flexibility for hardware constraints.

\begin{figure}[t]
\centering
\includegraphics[width=\linewidth]{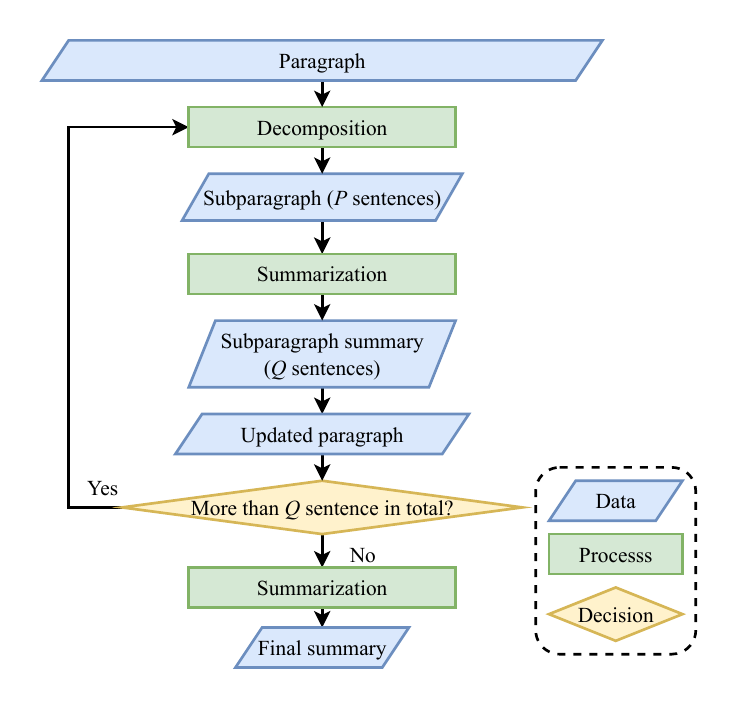}
\caption{Flowchart of decomposition in ES.}
\label{fig:ES_decomposition_workflow}
\end{figure}

To overcome this rigidity, we propose decomposing the ES task into a sequence of smaller subproblems as described in Figure~\ref{fig:ES_decomposition_workflow}. This multi-stage approach introduces intermediate steps that allow for controlled problem size. Specifically, while $N$ is fixed at the input stage and $M$ must be satisfied at the final output, intermediate stages permit tuning of both $N$ and $M$ within bounded ranges.


In the decomposition process, we select a subparagraph of $P$ consecutive sentences and summarize them into $Q$ sentences using the Ising formulation and COBI solver. For the first decomposition, selection begins at the start of the paragraph; for subsequent decompositions, it resumes from the position following the most recent decomposition. If the selection reaches the end of the paragraph, it wraps around to the beginning. After each subparagraph summarization, the original $P$ sentences are replaced with the generated $Q$-sentence summary. We then check whether the updated paragraph still contains more than $P$ sentences. If so, the decomposition process is repeated; otherwise, a final summarization is performed on the remaining paragraph to produce the $M$-sentence summary as the final ES output.

This decomposition strategy offers flexibility in shaping the distributions of $h_i$ and $J_{ij}$ across subproblems. By tuning $P$ and $Q$ and controlling the local problem context, we can better balance the magnitudes of local fields and coupling weights, improving numerical conditioning and enhancing solution quality on hardware-constrained Ising solvers such as COBI.

We assess the effectiveness of this approach on 20 CNN/DailyMail benchmarks, each containing $N=20$ sentences with a final target summary length of $M = 6$. To keep the setup simple while balancing coupling weight scales with the precision limits of COBI, we set $P = 20$ and $Q = 10$ in the decomposition workflow shown in Figure~\ref{fig:ES_decomposition_workflow}. Under this configuration, each 20-sentence paragraph is first summarized into a 10-sentence intermediate summary, which is then summarized into the final 6-sentence output. All summarizations use the improved Ising formulation in Eq.~\eqref{eq:es_ising_final}, with COBI solving each subproblem. The complete process thus involves solving two Ising instances per benchmark, one for the 20-sentence original paragraph to a 10-sentence intermediate summary and one for the final selection. To assess the performance of the decomposition approach, we compute the normalized objective value using~\eqref{eq:normalized_obj}, which compares the extracted summary against the ground-truth optimal solution from Gurobi~\cite{gurobi}.

\begin{figure}[t]
\centering
\includegraphics[width=0.8\linewidth]{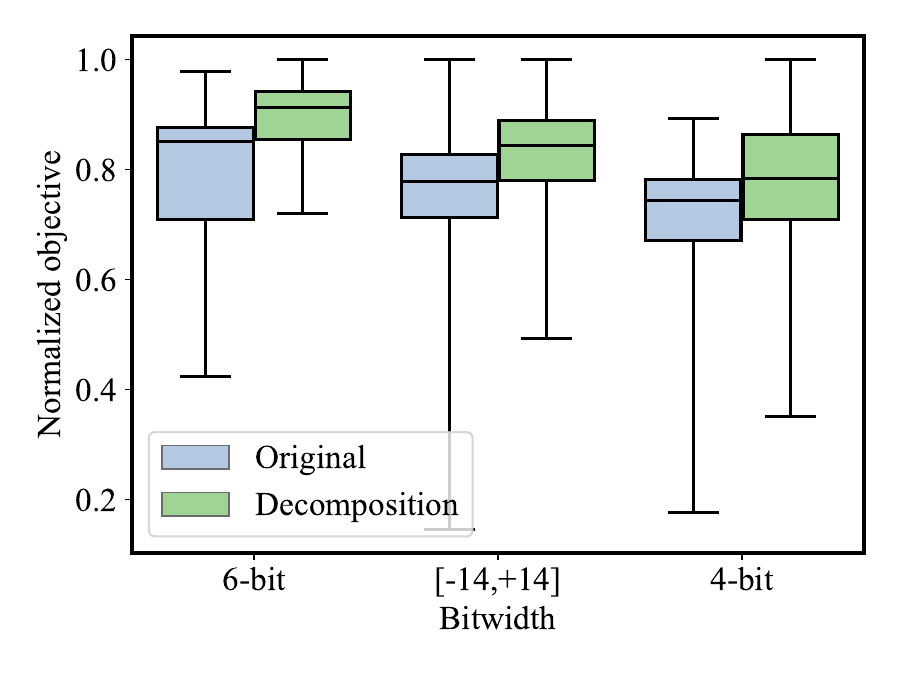}
\caption{Normalized objective for decomposition with various precision.}
\label{fig:ES_decompose}
\end{figure}

To keep stable, we repeat the entire process 100 times for each benchmark under various precision settings, ranging from 4-bit to 8-bit, including the integer range $[-14, +14]$ native to COBI. All Ising formulations are solved by Tabu search~\cite{misevicius2005tabu} as a simulation of COBI. The average normalized objective across repetitions is reported in Figure~\ref{fig:ES_decompose}. Each boxplot represents the distribution of normalized objectives: Blue plots correspond to the decomposition-based workflow, while orange plots represent the baseline approach that directly solves the full $N=20$, $M=6$ problem as a single Ising instance.

As shown in Figure~\ref{fig:ES_decompose}, the decomposition strategy consistently outperforms the direct formulation across all precision levels. In particular, for the COBI-compatible $[-14, +14]$ integer setting, the median normalized objective improves from 0.75 to 0.83. Although worst-case performance still leaves room for improvement, we note that iterative techniques such as stochastic rounding (see Section~\ref{ssec:ES_decomposition}) can further enhance precision and solution quality. Therefore, all subsequent hardware-based experiments adopt both decomposition and iterative stochastic refinement to fully exploit the capabilities of COBI on ES problems.

%% file: sec/5_Experiment.tex
\section{Hardware experiments}
\label{sec:ES_hardware_experiments}
After exploring the formulation strategies in Section~\ref{sec:ES_formulation} and the iterative and decomposition techniques introduced in Section~\ref{sec:ES_workflow} using a Tabu sampler, we integrate all these components into a unified workflow and evaluate its performance on the COBI hardware~\cite{cilasun2025coupled}. The complete workflow is based on the decomposition workflow; each benchmark (i.e., a paragraph with many sentences) is decomposed into smaller subparagraphs. For each subproblem, we generate an optimized Ising formulation in floating-point precision based on our improved model. We then apply stochastic rounding to quantize this formulation into the $[-14, +14]$ integer range supported by COBI, producing a hardware-compatible Ising instance. COBI is used iteratively to solve these instances, and the selected sentences from each subproblem are retained.

To assess accuracy, we use Gurobi to compute the ground-truth maximal and minimal objective values for each benchmark using the formulation in~\eqref{eq:es_opt_mu_beta}. The quality of our results is evaluated using the normalized objective metric defined in~\eqref{eq:normalized_obj} based on the maximum and minimum. To ensure statistical stability, each benchmark is solved independently 100 times, and we report the average normalized objective across these repetitions for each benchmark. 

We evaluate the proposed workflow on the CNN/DailyMail dataset using the same 20 benchmarks of 20-sentence paragraphs ($N=20$) employed in the Tabu simulations described in Section~\ref{sec:ES_workflow}. In addition, we include 20 benchmarks of 50-sentence paragraphs from the CNN/DailyMail dataset and 20 benchmarks of 100-sentence paragraphs from the Extreme Summarization (XSum) dataset~\cite{narayan2018don}. Each paragraph is summarized into $M=6$ sentences. The decomposition parameters are fixed to $P=20$ and $Q=10$, consistent with Section~\ref{ssec:ES_decomposition}, to satisfy the precision constraints of the COBI hardware.


\begin{figure}[t]
    \centering
    \begin{subfigure}[b]{0.5\linewidth}
        \centering
        \includegraphics[width=\linewidth]{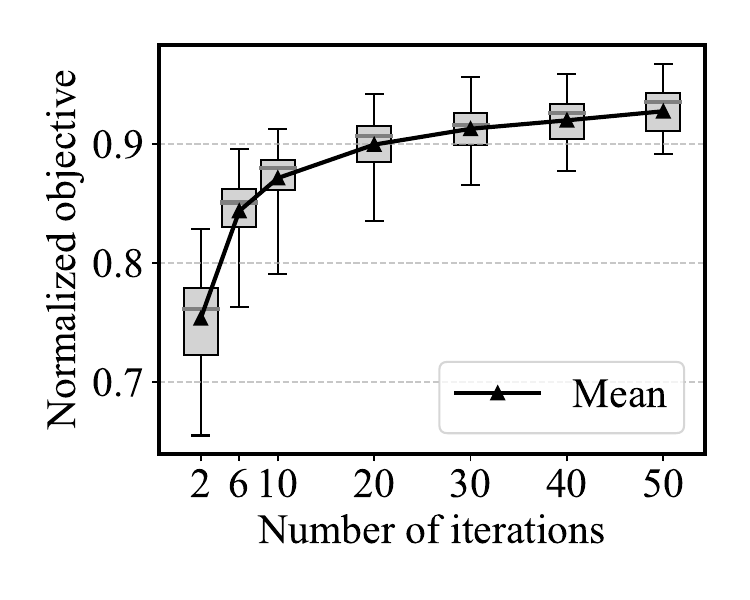}
        \label{fig:ES_cobi_50}
        \vspace{-0.3in}
        \caption{}
    \end{subfigure}
    \hspace{-0.02\linewidth}
    \begin{subfigure}[b]{0.5\linewidth}
        \centering
        \includegraphics[width=\linewidth]{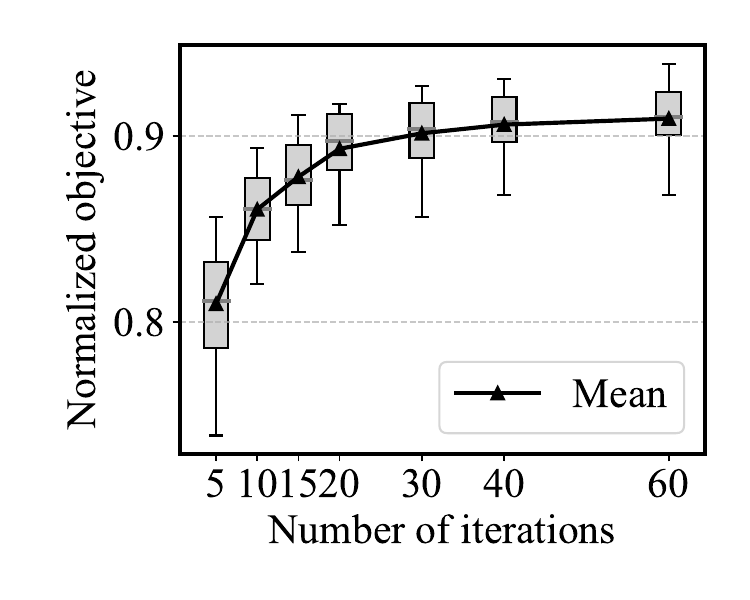}
        \label{fig:ES_cobi_20}
        \vspace{-0.3in}
        \caption{}
    \end{subfigure}
    \begin{subfigure}[b]{0.5\linewidth}
        \centering
        \includegraphics[width=\linewidth]{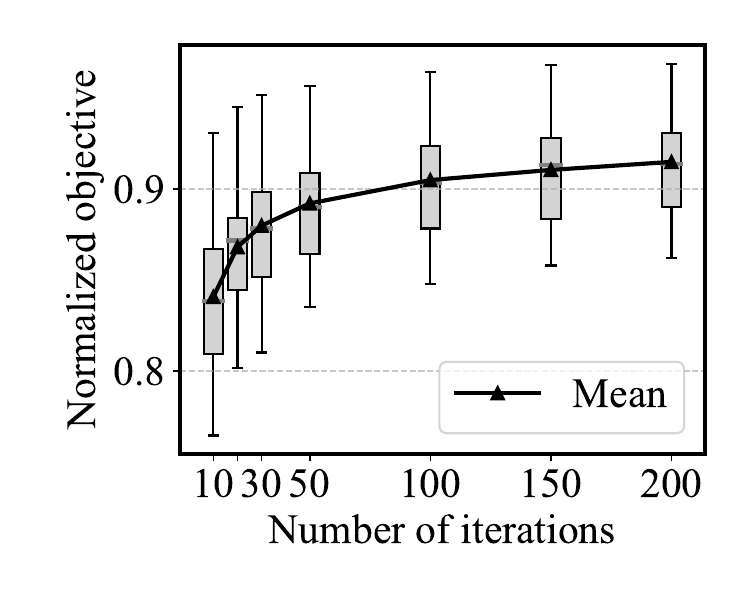}
        \label{fig:ES_cobi_100}
        \vspace{-0.3in}
        \caption{}
    \end{subfigure}
    \hspace{-0.02\linewidth}
    \begin{subfigure}[b]{0.5\linewidth}
        \centering
        \includegraphics[width=\linewidth]{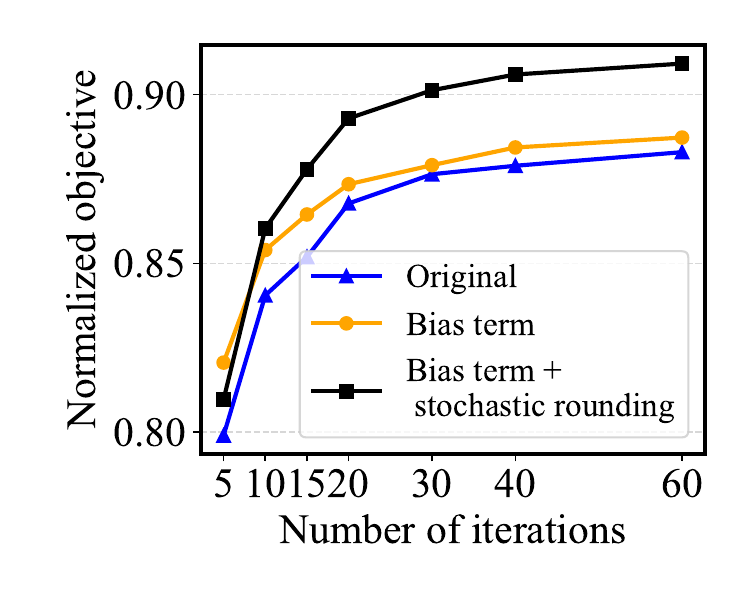}
        \label{fig:ES_cobi_ablation}
        \vspace{-0.3in}
        \caption{}
    \end{subfigure}
    \caption{Comparison of COBI accuracy on (a) 20-sentence benchmarks, (b) 50-sentence benchmarks, and (c) 100-sentence benchmarks across different numbers of iterations, (d) ablation study on 50-sentence benchmarks.}
    \label{fig:ES_cobi_results}
\end{figure}

Figure~\ref{fig:ES_cobi_results}(a), (b) and (c) compares the accuracy of COBI, the Tabu search solver (under the same precision as COBI), and the random baseline described in Section~\ref{ssec:ES_iterations}. The x-axis indicates the total number of iterations~(defined in Section~\ref{ssec:ES_iterations}), while the y-axis shows the normalized objective. The box plot summarizes the accuracy of COBI across 20 benchmarks: whiskers show the minimum and maximum values, boxes represent the interquartile range with medians indicated, and the black line marks the average score. 

Figure~\ref{fig:ES_cobi_results}(d) presents an ablation study on the 50-sentence benchmarks, examining the effects of the bias term~(mentioned in Section~\ref{ssec:ES_Formulation_improved}) and stochastic rounding~(in Section~\ref{ssec:ES_iterations}). Each data point in Figure~\ref{fig:ES_cobi_results}(d) represents the average normalized objective over 100 independent runs across 20 benchmarks. The results show that incorporating the bias term consistently improves performance compared to the original baseline. In addition, stochastic rounding further enhances solution quality as the number of iterations increases. With a small number of iterations, performance may be unstable due to the randomness introduced by stochastic rounding; however, iterative refinement mitigates this variability and leads to higher and more reliable accuracy.


As defined in Section~\ref{ssec:ES_iterations}, each iteration corresponds to solving a single Ising formulation on either COBI or Tabu. In the decomposition workflow, obtaining a final solution requires solving at least two Ising subproblems for 20-sentence benchmarks, so all decomposition-based data points have an even iteration count. The number of stochastic rounding iterations is therefore half the total iteration count. For example, the configuration with 10 total iterations corresponds to two decomposition steps, each with its Ising formulation solved with 5 stochastic rounding iterations.

Overall, COBI attains slightly lower average accuracy than Tabu but consistently outperforms the random baseline for all iteration counts above two. The gap relative to Tabu is likely due to the higher output variability of COBI, whereas the software-based Tabu sampler delivers consistently high accuracy on integer-weight Ising formulations. At two iterations, the accuracy of COBI drops sharply—likely a result of instability from combining decomposition with stochastic rounding at such a low iteration count, amplified by the inherent variability of COBI hardware. As the iteration count increases, the accuracy of COBI improves substantially and converges toward that of Tabu. Specifically, under 50 iterations, COBI can achieve an average of 92.8\% of the normalized objective, which is very close to the Tabu results, 93.5\%.

Despite a modest accuracy difference, the hardware advantages of COBI are significant: with a runtime of approximately \(200\,\mu\text{s}\) and a power consumption of only 24~mW, it provides substantial improvements in both speed and energy efficiency compared to Tabu, which typically requires 25~ms on a CPU consuming 20~W. These results demonstrate that CMOS-based COBI hardware is a practical and efficient accelerator for extractive summarization, even under stringent precision requirements. Projected runtimes and energy consumption for COBI are reported in Table~\ref{tbl:ES_cobi_runtime_energy_20}. Compared to the Tabu solver on CPU, which typically demands 25~ms runtime and 500~mJ of energy, COBI achieves 1-2 orders of magnitude faster runtime and 2-3 orders of magnitude lower energy consumption across the normalized objective range of 0.8 to 0.92.

\begin{table}[t]
\centering
\begin{tabular}{|c|c|c|c|}
\hline
\makecell{Normalized \\ objective} & \makecell{Number of \\ iterations} & Runtime (ms) & Energy (J) \\ \hline
0.8 & 4.06 & 1.62 & 0.390 \\ \hline
0.85 & 7.10 & 2.84 & 0.682 \\ \hline
0.9 & 19.62 & 7.85 & 0.188 \\ \hline
0.91 & 24.15 & 9.66 & 0.232 \\ \hline
0.92 & 29.15 & 11.66 & 0.280 \\ \hline
\end{tabular}
\caption{Projected COBI runtime and energy under various normalized objectives.}
\label{tbl:ES_cobi_runtime_energy_20}
\end{table}

In addition to the 20 benchmarks containing 20 sentences each, we consider 20 additional benchmarks with 50 sentences from the CNN/DailyMail dataset. For all benchmarks, the task is to generate a six-sentence extractive summary using the decomposition workflow illustrated in Figure~\ref{fig:ES_decomposition_workflow}, with COBI employed as the Ising solver.

To evaluate and compare the runtime and energy efficiency of different solvers, we use the Time to Solution~(TTS) and Energy to Solution~(ETS) metrics. TTS and ETS are defined as the total runtime and energy required to achieve a probability \(p_{\text{target}}\) of obtaining a normalized objective of at least 0.9. The threshold of 0.9 serves as a practical benchmark for both solution quality and solver efficiency. An illustrative example is provided in the Supplementary Materials, where we show two six-sentence summaries generated from the same paragraph: one with the optimal objective and another with a 0.9 normalized objective. The comparison demonstrates that the summary with a 0.9 normalized objective conveys information very close to the optimal one from a reader's perspective. Based on this observation, we treat summaries with normalized objective values of at least 0.9 as successful solutions in our experiments.

The TTS is estimated using a maximum likelihood estimation~(MLE)-based approach, where each solver run is treated as a Bernoulli trial and a successful trial is defined as achieving a normalized objective of at least 0.9. Using the empirical relationship between the number of iterations and the normalized objective, similar to Figure~\ref{fig:ES_iterations_20}, we determine the iteration count at which the success threshold is first reached. Let \(k_i\) denote this iteration count for benchmark \(i\). Assuming a constant probability of success per iteration, the likelihood of observing the set \(\{k_i\}\) is modeled as a geometric process. The success probability is then estimated as
\begin{equation}
\hat{p}_{\text{success}} = \frac{1}{\hat{k}}, \quad \text{where} \quad \hat{k} = \frac{1}{n} \sum_{i=1}^{n} k_i,
\end{equation}
which corresponds to the MLE of the mean number of iterations required to reach a normalized objective of 0.9.

The TTS for COBI, brute-force, and Tabu is subsequently computed as
\begin{equation}
\text{TTS} = \frac{\ln(1-p_{\text{target}})}{\ln(1-\hat{p}_{\text{success}})} \cdot \frac{1}{\nu} \sum_{i=1}^{\nu} \text{Runtime}_i,
\end{equation}
where \(p_{\text{target}} = 0.95\) is the desired success probability, \(\nu\) denotes the number of independent runs, and \(\text{Runtime}_i\) is the runtime of the \(i^\text{th}\) run. Since stochastic rounding is applied in each iteration, the TTS calculation includes the objective evaluation time of \(18.9~\mu\mathrm{s}\) per iteration.

Based on the TTS, the ETS is defined as
\begin{equation}
\text{ETS} = \text{TTS}_{\text{COBI}} \cdot P_{\text{COBI}} + \text{TTS}_{\text{software}} \cdot P_{\text{CPU}},
\end{equation}
where \(P_{\text{COBI}}\) and \(P_{\text{CPU}}\) denote the power dissipation of the COBI solver and the CPU, respectively. For COBI, \(\text{TTS}_{\text{software}}\) corresponds to the objective evaluation time of \(18.9~\mu\mathrm{s}\)  under stochastic rounding. In the case of brute-force and Tabu implementations, \(\text{TTS}_{\text{COBI}}\) is not applicable, and the ETS reduces to \(\text{TTS}_{\text{software}} \cdot P_{\text{CPU}}\). In our evaluation, the power of the CPU is assumed to be 20 watts, while the power of COBI is 25 mW. The CPU power consumption corresponding to the objective evaluation time is incorporated into the ETS computation.

\begin{figure}[t]
    \centering
    \begin{subfigure}[b]{0.5\linewidth}
        \centering
        \includegraphics[width=\linewidth]{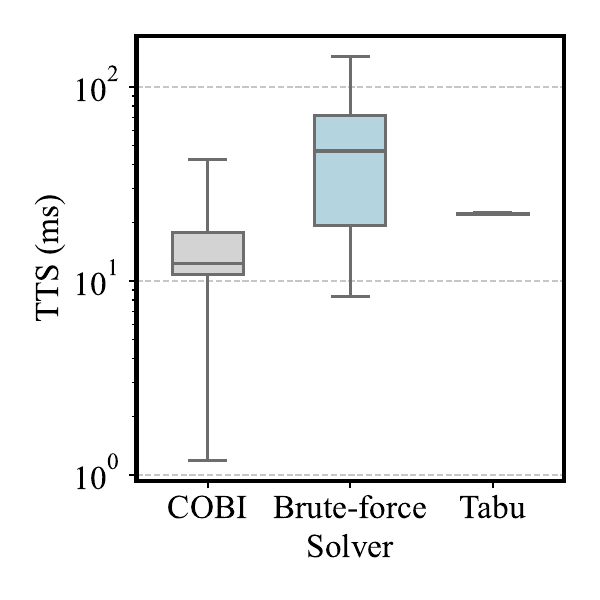}
        \vspace{-0.3in}
        \caption{}
        \label{fig:ES_tts_20}
    \end{subfigure}
    \hspace{-0.02\linewidth} 
    \begin{subfigure}[b]{0.5\linewidth}
        \centering
        \includegraphics[width=\linewidth]{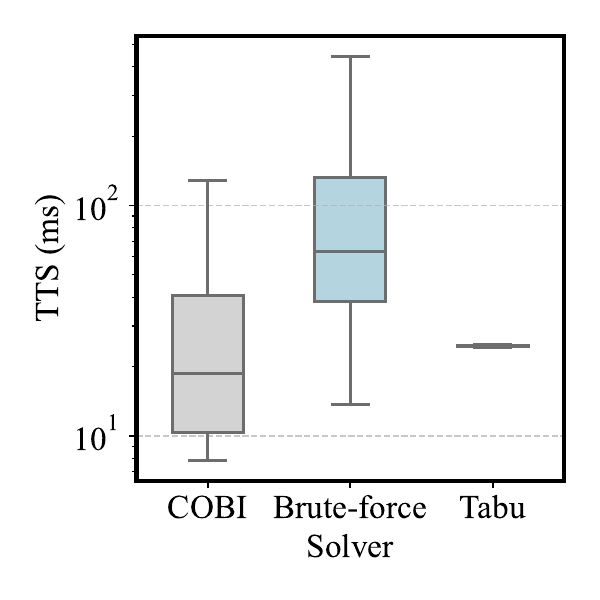}
        \vspace{-0.3in}
        \caption{}
        \label{fig:ES_tts_50} 
    \end{subfigure}
        \begin{subfigure}[b]{0.5\linewidth}
        \centering
        \includegraphics[width=\linewidth]{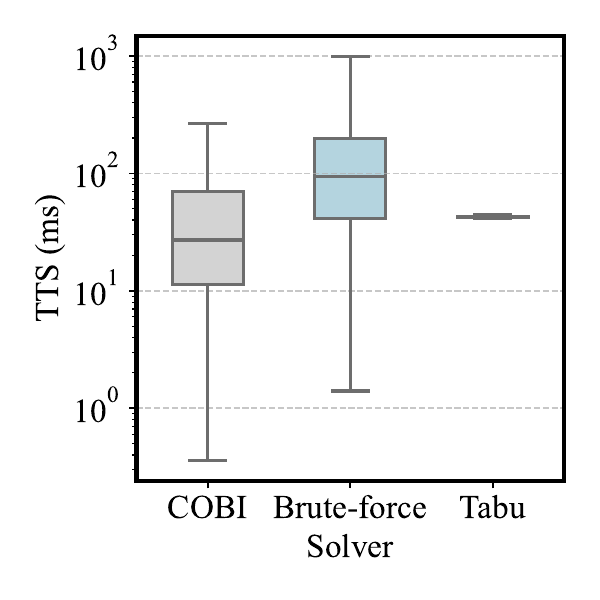} 
        \vspace{-0.3in}
        \caption{}
        \label{fig:ES_tts_100} 
    \end{subfigure}
    \caption{Comparison of TTS on COBI, brute-force, and Tabu on (a) 20-sentence benchmarks, (b) 50-sentence benchmarks, and (c) 100-sentence benchmarks across different numbers of iterations.}
    \label{fig:ES_tts_results}
\end{figure}

By estimating $p_{\text{success}}$ from the iteration–objective curves via MLE and substituting it into the above equation, we obtain a direct, data-driven estimate of the TTS for each solver.

\begin{figure}[t]
    \centering
    \begin{subfigure}[b]{0.5\linewidth}
        \centering
        \includegraphics[width=\linewidth]{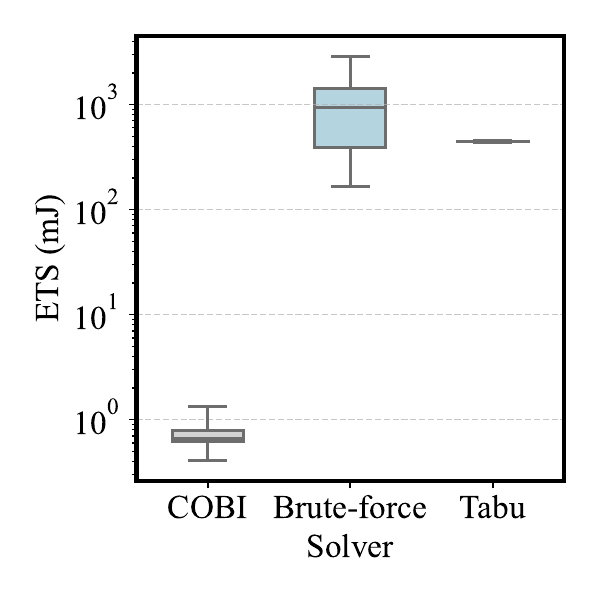}
        \label{fig:ES_ets_20}
        \vspace{-0.3in}
        \caption{}
    \end{subfigure}
    \hspace{-0.02\linewidth} 
    \begin{subfigure}[b]{0.5\linewidth}
        \centering
        \includegraphics[width=\linewidth]{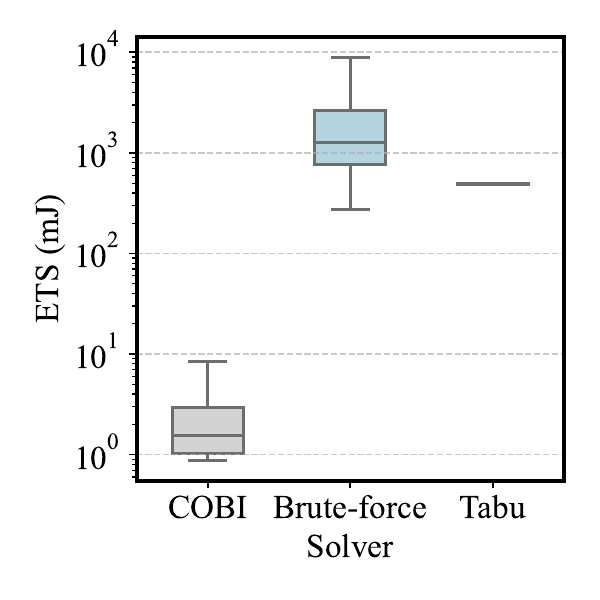}
        \label{fig:ES_ets_50}
        \vspace{-0.3in}
        \caption{}
    \end{subfigure}
    \begin{subfigure}[b]{0.5\linewidth}
        \centering
        \includegraphics[width=\linewidth]{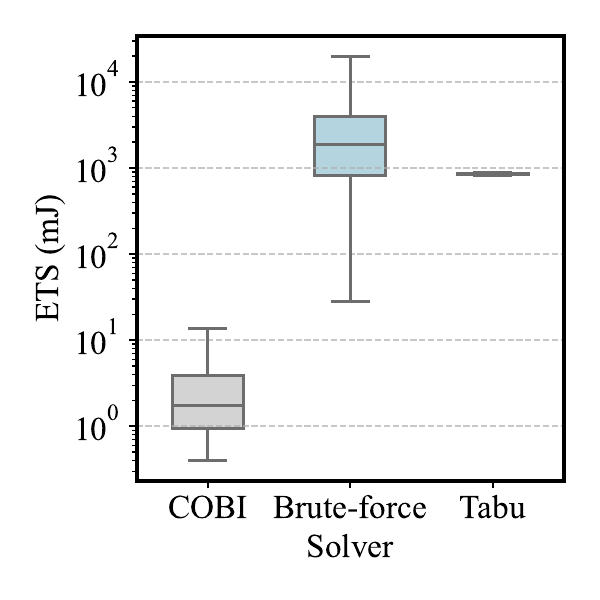} 
        \vspace{-0.3in}
        \caption{}
        \label{fig:ES_ets_100} 
    \end{subfigure}
    \caption{Comparison of ETS on COBI, brute-force, and Tabu on (a) 20-sentence benchmarks, (b) 50-sentence benchmarks, and (c) 100-sentence benchmarks across different numbers of iterations.}
    \label{fig:ES_ets_results}
\end{figure}

The TTS values for COBI, the brute-force baseline, and Tabu are shown in Figure~\ref{fig:ES_tts_results}. For the 20-sentence benchmarks, COBI achieves an average TTS of 16.6~ms, compared to 50.9~ms for brute-force and 22.2~ms for Tabu, corresponding to a $3.1\times$ speedup over brute-force and a $1.3\times$ speedup over Tabu. For the 50-sentence benchmarks, the average TTS for COBI is 29.4~ms, compared to 122.9~ms for brute-force giving a $4.2\times$ speedup over brute-force. For the 100-sentence benchmarks, the average TTS values are 56.5~ms for COBI, and 240.3~ms for brute-force, where COBI achieves a $4.3\times$ speedup over brute-force. 


Using the measured power consumption of 25~mW for COBI and a CPU power of 20~W, the ETS values are computed directly from the corresponding TTS measurements. For the COBI implementation with stochastic rounding (including objective evaluation), both the COBI runtime and the CPU runtime for evaluation are incorporated into the ETS calculation. The resulting ETS values for COBI, brute-force, and Tabu are shown in Figure~\ref{fig:ES_ets_results}. COBI achieves an energy reduction of 2--3 orders of magnitude compared to both Tabu and the brute-force baseline.

%% file: sec/6_Conclusion.tex
\section{Conclusion}
\label{sec:ES_conclusion}

This work presents a complete hardware-aware workflow for solving the extractive summarization problem on a CMOS-based Ising machine. The proposed approach demonstrates how high-density combinatorial optimization problems can be made compatible with low-precision Ising hardware by combining formulation-level and architectural techniques. Starting from an ILP-based formulation of extractive summarization, we derive an improved Ising model that balances local-field and coupling terms, and introduce stochastic rounding to bridge the gap between floating-point formulation and integer-valued hardware constraints. To handle larger inputs, we employ a decomposition strategy that partitions the original problem into smaller Ising subproblems that can be efficiently solved by the COBI hardware~\cite{Lo2023}. This unified workflow tightly integrates problem formulation, decomposition, and iterations with stochastic rounding to fully exploit the capabilities of the COBI architecture. Experimental results on the CNN/DailyMail dataset show that COBI achieves a 3.1--4.3$\times$ speedup over the brute-force baseline and comparable TTS to Tabu search, while delivering approximately three orders of magnitude lower ETS than brute-force and about 2.5 orders of magnitude lower ETS than Tabu, all while maintaining competitive summary quality (normalized objective $\ge 0.9$). These results highlight the potential of CMOS Ising solvers as low-energy hardware accelerators for real-time extractive summarization and, more broadly, for edge-deployable constrained optimization workloads.